\documentclass{article}

\usepackage{microtype}
\usepackage{graphicx}
\usepackage{subcaption}
\usepackage{booktabs} 

\usepackage{hyperref}

\usepackage[preprint]{icml2026}

\usepackage{amsmath}
\usepackage{amssymb}
\usepackage{mathtools}
\usepackage{amsthm}
\usepackage{multirow}
\usepackage{xcolor}

\usepackage[capitalize,noabbrev]{cleveref}

\definecolor{angularcolor}{HTML}{E63946}
\definecolor{magnitudecolor}{HTML}{457B9D}

\icmltitlerunning{Disentangling Direction and Magnitude in Transformer Representations}

\begin{document}

\twocolumn[
  \icmltitle{Disentangling Direction and Magnitude in Transformer Representations:\\
    A Double Dissociation Through L2-Matched Perturbation Analysis}


\begin{icmlauthorlist}
  \icmlauthor{Mangadoddi Srikar Vardhan}{inst1}
  \icmlauthor{Lekkala Sai Teja}{inst2}
\end{icmlauthorlist}

\icmlaffiliation{inst1}{NIT Silchar, India}
\icmlaffiliation{inst2}{NIT Silchar, India}

\icmlcorrespondingauthor{Mangadoddi Srikar Vardhan}{mangadoddis\_ug\_22@cse.nits.ac.in}
  \icmlkeywords{Machine Learning, Transformers, Interpretability, Representation Learning, Neural Networks}

  \vskip 0.3in
]

\printAffiliationsAndNotice{\icmlEqualContribution \textit{Preprint. Under review.}} 

\begin{abstract}
Transformer hidden states encode information as high-dimensional vectors, yet whether \textit{direction} (orientation in representational space) and \textit{magnitude} (vector norm) serve distinct functional roles remains unclear. Studying Pythia-family models, we discover a striking \textbf{cross-over dissociation}: angular perturbations cause up to 42.9$\times$ more damage to language modeling loss, while magnitude perturbations cause disproportionately more damage to syntactic processing (20.4\% vs.\ 1.6\% accuracy drop on subject-verb agreement). This finding is enabled by \textbf{L2-matched perturbation analysis}, a methodology ensuring that angular and magnitude perturbations achieve identical Euclidean displacements. Causal intervention reveals that angular damage flows substantially through the attention pathways (28.4\% loss recovery via attention repair), while magnitude damage flows partly through the LayerNorm pathways (29.9\% recovery via LayerNorm repair). These patterns replicate across scales within the Pythia architecture family.These findings provide evidence that direction and magnitude support partially distinct computational roles in LayerNorm-based architectures. The direction preferentially affects attentional routing, while magnitude modulates processing intensity for fine-grained syntactic judgments. We find different patterns in RMSNorm-based architectures, suggesting that the dissociation depends on architectural choices. Our results refine the linear representation hypothesis and have implications for model editing and interpretability research.
\end{abstract}

\section{Introduction}

The question of how neural language models encode and manipulate information has become central to both interpretability research and the broader scientific understanding of artificial intelligence. Transformer architectures \citep{vaswani2017attention} represent linguistic information as high-dimensional vectors that flow through successive layers of attention and feed-forward computation. Yet, a fundamental question remains underexplored: \textit{what aspects of these vectors actually matter for downstream computation?}

Every vector in a transformer's hidden state can be decomposed into two geometric properties: its \textbf{direction} (the orientation in representational space, typically represented as a unit vector) and its \textbf{magnitude} (the L2 norm, representing the vector's length). The influential \textit{linear representation hypothesis} \citep{park2023linear, nanda2023emergent, elhage2022toy} posits that transformers encode semantic features as directions in the activation space specific concepts correspond to specific orientations, and the presence of a feature can be detected by projection onto its associated direction. This hypothesis has proven to be exceptionally productive, enabling techniques ranging from activation patching \citep{meng2022locating, wang2022interpretability} to linear probing \citep{belinkov2017neural, hewitt2019structural}.

However, the linear representation hypothesis, as typically formulated, is silent on the role of magnitude. This silence is consequential. Vector norms in transformer representations are not constant; they vary substantially across tokens, layers, and contexts \citep{kobayashi2020attention, ethayarajh2019contextual}. Several lines of evidence suggest that magnitude carries meaningful information: LayerNorm operations explicitly manipulate norms \citep{ba2016layer, xu2019understanding}, attention mechanisms compute over both direction and magnitude \citep{elhage2021mathematical}, and recent work on representation engineering suggests that scaling vectors can modulate behavioral tendencies \citep{turner2023activation, zou2023representation}. However, no systematic framework exists for disentangling the functional contributions of these two geometric properties.

\textbf{Our Central Finding.}Studying Pythia-family models (410M-1.4B parameters) \citep{biderman2023pythia}, we discover a striking pattern that replicates within this architecture family but shows different behavior in other architectures:

\textbf{Crossover Dissociation:} Angular perturbations cause dramatically more damage to language modeling loss (up to 42.9$\times$ at small displacements), while magnitude perturbations cause more damage to syntactic accuracy (up to 21.1\% greater drop on BLiMP subject-verb agreement at $\delta=10$).

This Crossover dissociation where each perturbation type preferentially impairs a different cognitive function—provides evidence that direction and magnitude support \textit{partially distinct computational roles} in LayerNorm based transformers. In the terminology of cognitive neuroscience, this pattern suggests a form of \textit{partial functional specialization} \citep{shallice1988neuropsychology, teuber1955physiological}.

\textbf{The Confounding Problem and Our Solution.} A naive approach to comparing direction and magnitude would simply perturb each independently and measure downstream effects. But this comparison is fundamentally confounded: a small angular rotation and a small magnitude scaling produce perturbations of vastly different \textit{sizes} in representational space. If angular perturbations cause more damage, is this because direction is more important, or simply because the angular perturbation was larger? Without controlling for perturbation size, no fair comparison is possible.

Our finding is enabled by \textbf{L2-matched perturbation analysis}, a methodology that addresses this confound by ensuring both angular and magnitude perturbations achieve \textit{identical Euclidean displacements} $\delta$ from the original representation. This matching enables, for the first time, a controlled comparison under matched Euclidean displacement of the functional importance of direction versus magnitude in transformer computation.

\textbf{Mechanistic Analysis.}  We further investigate the mechanistic basis of this dissociation through causal intervention experiments. Our analysis reveals that angular damage flows substantially through attention pathways: repairing attention patterns recovers 28.4\% of angular-induced loss versus only 15.2\% for magnitude perturbations in Pythia-410M, a qualitative pattern that replicates in Pythia-1.4B. Magnitude damage, in contrast, flows partly through LayerNorm pathways: repairing LayerNorm outputs recovers 29.9\% of magnitude-induced damage versus only 13.7\% for angular perturbations. This suggests a model in which \textbf{direction preferentially affects attentional routing} which tokens should attend to which—while \textbf{magnitude modulates processing intensity} through LayerNorm-mediated pathways. We note that these pathways account for roughly 30\% of each effect; the complete mechanistic picture remains an area for future work. While these pathways account for a significant portion of the observed effects, the complete mechanistic picture remains an area for future investigation.

Our contributions are:
\begin{enumerate}
    \item \textbf{Empirical:} Discovery of a cross-over dissociation between angular perturbations (loss-critical) and magnitude perturbations (syntax-critical) in Pythia-family models.
    \item \textbf{Methodological:} L2-matched perturbation analysis, enabling controlled comparison of direction vs.\ magnitude importance.
    \item \textbf{Mechanistic:} Causal evidence that angular effects flow substantially through attention pathways while magnitude effects flow partly through LayerNorm pathways, with the caveat that these pathways explain approximately 30\% of each effect.
    \item \textbf{Theoretical:} Refinement of the linear representation hypothesis to incorporate partial functional specialization of geometric properties.
\end{enumerate}
\section{Related Work}

\textbf{The Linear Representation Hypothesis.} A growing body of work suggests that transformers encode concepts as linear directions in activation space. \citet{mikolov2013linguistic} first demonstrated linear regularities in word embeddings; subsequent work extended this to contextual representations \citep{ethayarajh2019contextual}, finding that contextualized embeddings occupy anisotropic subspaces. \citet{park2023linear} and \citet{nanda2023emergent} formalized the linear representation hypothesis, showing that features from sentiment to factual knowledge can be extracted via linear probes. \citet{elhage2022toy} demonstrated that even toy models learn interpretable linear features. Our work extends this hypothesis by providing evidence that direction and magnitude serve partially distinct computational roles.

\textbf{Representation Geometry.} The geometric structure of transformer representations has received substantial attention. \citet{ethayarajh2019contextual} showed that contextual embeddings are highly anisotropic, with most variance concentrated in a low dimensional subspace. \citet{timkey2021all} examined how this geometry evolves across layers. \citet{cai2021isotropy} and \citet{bis2021too} investigated the relationship between isotropy and model performance. \citet{kobayashi2020attention} analyzed how vector norms interact with attention weights. Our perturbation methodology provides a causal complement to these correlational analyses, directly testing whether geometric properties are functionally necessary.

\textbf{Activation Patching and Causal Intervention.} Our methodology builds on the tradition of causal intervention in neural networks. \citet{vig2020investigating} introduced activation patching for studying gender bias; \citet{meng2022locating} extended this to factual knowledge localization. \citet{wang2022interpretability} used patching to study indirect object identification. \citet{geiger2021causal} formalized causal abstraction for neural networks. Unlike previous work that patches entire activations, we \textit{decompose} activations into direction and magnitude, enabling finer-grained causal analysis.

\textbf{Attention Mechanisms.} The role of attention in transformer computation has been extensively studied. \citet{clark2019does} and \citet{kovaleva2019revealing} characterized attention patterns in BERT. \citet{elhage2021mathematical} provided a mathematical framework for understanding attention as information movement. \citet{olsson2022context} identified induction heads as key computational motifs. Our finding that angular perturbations preferentially disrupt attention patterns connects geometric representation to mechanistic function.
\section{L2 Matched Perturbation Analysis}

The core methodological contribution of this paper is a framework for comparing the functional importance of direction versus magnitude while controlling for perturbation size. We first formalize the problem, then describe our solution.
\subsection{Problem Formulation}

Let $\mathbf{h} \in \mathbb{R}^d$ be a hidden state vector in a transformer with $d$ dimensions. We can decompose $\mathbf{h}$ into:
\begin{align}
    \mathbf{h} &= \|\mathbf{h}\| \cdot \hat{\mathbf{h}}
\end{align}
where $\|\mathbf{h}\|$ is the L2 norm (magnitude) and $\hat{\mathbf{h}} = \mathbf{h}/\|\mathbf{h}\|$ is the unit direction vector.

A \textbf{magnitude perturbation} scales the norm while preserving direction:
\begin{align}
    \mathbf{h}'_{\text{mag}} &= \alpha \mathbf{h} = (\alpha \|\mathbf{h}\|) \cdot \hat{\mathbf{h}}
\end{align}

An \textbf{angular perturbation} rotates the direction while preserving norm:
\begin{align}
    \mathbf{h}'_{\text{ang}} &= \|\mathbf{h}\| \cdot \hat{\mathbf{h}}'
\end{align}
where $\hat{\mathbf{h}}'$ is a unit vector rotated from $\hat{\mathbf{h}}$ by angle $\theta$ in a randomly sampled orthogonal direction, serving as a controlled geometric perturbation.

The \textbf{confounding problem} is that these perturbations have different Euclidean distances from the original. For a magnitude perturbation with scaling factor $\alpha$ and an angular perturbation with rotation angle $\theta$:
\begin{align}
    \|\mathbf{h} - \mathbf{h}'_{\text{mag}}\| &= |1-\alpha| \cdot \|\mathbf{h}\| \\
    \|\mathbf{h} - \mathbf{h}'_{\text{ang}}\| &= \|\mathbf{h}\| \cdot \sqrt{2(1-\cos\theta)}
\end{align}

Without matching these distances, any comparison is confounded by perturbation size alone. For reference, hidden state norms in Pythia-410M (in the layers we intervene on) typically range from 15-25, meaning a displacement of $\delta = 5$ represents roughly 20-30\% of the typical vector magnitude. We return to the potential anisotropy of representation space in Section~\ref{sec:limitations}.
\subsection{L2-Matched Perturbations}

Our solution is to constrain both perturbations to achieve the same Euclidean displacement $\delta$:
\begin{align}
    \|\mathbf{h} - \mathbf{h}'_{\text{mag}}\| = \|\mathbf{h} - \mathbf{h}'_{\text{ang}}\| = \delta
\end{align}

\textbf{Magnitude Perturbation.} Given target displacement $\delta$, we compute the scaling factor from $|1-\alpha|\cdot\|\mathbf{h}\| = \delta$:
\begin{align}
    \alpha = 1 \pm \frac{\delta}{\|\mathbf{h}\|}
\end{align}
We randomly select the plus or minus branch (scaling up or down) for each perturbation and report results averaged across both; this requires $\delta < \|\mathbf{h}\|$. This constraint is satisfied in all our experiments since typical hidden state norms at our intervention layers are 15-25 and our maximum $\delta = 20$. This yields $\mathbf{h}'_{\text{mag}} = \alpha \mathbf{h}$ with $\|\mathbf{h} - \mathbf{h}'_{\text{mag}}\| = \delta$.

\textbf{Angular Perturbation.} We seek a rotation achieving displacement $\delta$ while preserving norm. We sample a random orthogonal direction $\mathbf{v} \perp \mathbf{h}$, treating this as a controlled geometric probe of direction, then compute the rotation angle $\theta$ in closed form:
\begin{align}
    \mathbf{h}'_{\text{ang}} &= \|\mathbf{h}\| \cdot \left( \cos\theta \cdot \hat{\mathbf{h}} + \sin\theta \cdot \hat{\mathbf{v}} \right)
\end{align}
where $\theta$ is derived from the constraint $\|\mathbf{h} - \mathbf{h}'_{\text{ang}}\| = \delta$:
\begin{align}
    \theta = \arccos\left(1 - \frac{\delta^2}{2\|\mathbf{h}\|^2}\right)
\end{align}
This formula requires $\delta \leq 2\|\mathbf{h}\|$, which is satisfied throughout our experiments.

\textbf{Verification.} We empirically verify that all perturbations achieve the target $\delta$ within tolerance 0.01 at the intervention layers across all experiments. This ensures that differences in downstream effects cannot be attributed to initial perturbation size alone and instead reflect differences in perturbation type. We analyze how these matched perturbations propagate through subsequent layers in Section~\ref{sec:results}.

\subsection{Experimental Protocol}

We apply perturbations to hidden states at intervention layers $\ell \in \mathcal{L}$ during the forward pass:
\begin{align}
    \mathbf{h}^{(\ell)} \leftarrow \text{Perturb}(\mathbf{h}^{(\ell)}, \delta, \text{type})
\end{align}
where type $\in$ \{angular, magnitude\}. We target middle layers ($\mathcal{L} = \{8, \ldots, 15\}$), where prior work suggests both syntactic and semantic processing are most active \citep{tenney2019bert, jawahar2019does}. We measure downstream effects on:
\begin{itemize}
    \item \textbf{Language modeling loss:} Cross-entropy on next-token prediction
    \item \textbf{Syntactic accuracy:} Performance on BLiMP \citep{warstadt2020blimp} subject-verb agreement minimal pairs
\end{itemize}

All experiments use 5 random seeds, each with independently sampled random orthogonal perturbation directions. We report means and standard errors across seeds, and assess statistical significance using paired t-tests with Bonferroni correction; we acknowledge that statistical power is limited by the small number of seeds ($n=5$), though effect sizes are large.

We conduct extensive layer-wise analyses: examining how perturbations propagate through subsequent layers, which layers contribute most to mechanistic recovery, and whether syntactic encoding varies by depth (Sections~\ref{sec:results} and \ref{sec:mechanistic}). Separately, we conduct causal intervention experiments (attention repair, LayerNorm repair) to investigate mechanistic pathways; these are described in Section~\ref{sec:mechanistic}.

\section{Experimental Setup}

\subsection{Models}
Our primary experiments use \textbf{Pythia-410M} \citep{biderman2023pythia}, a 24-layer transformer with 1024-dimensional hidden states and 16 attention heads, run in float32 precision for numerical stability during interventions. We choose Pythia for its well-documented training procedure and public availability of intermediate checkpoints. We replicate key findings on \textbf{Pythia-1.4B} (24 layers, 2048 dimensions) using the same data and protocol to assess scale generalization.

\subsection{Data}
\textbf{Language Modeling.} We evaluate on 281 sentences from WikiText-103 \citep{merity2016pointer}, filtered by length (30--300 characters) and sentence structure, yielding sequences of 10-64 tokens (mean 42.6). Baseline perplexity on this subset is 60.7 (loss $4.107 \pm 0.800$).

\textbf{Syntactic Evaluation.} We use the subject-verb agreement subset of BLiMP \citep{warstadt2020blimp}, combining \texttt{irregular\_plural\_subject\_verb\_agreement} and \texttt{regular\_plural\_subject\_verb\_agreement} paradigms (200 minimal pairs total). These items test whether the model assigns higher probability to grammatical sentences (e.g., ``The dogs \textit{run}'') versus ungrammatical alternatives (``The dogs \textit{runs}''). Baseline accuracy is 89.5\%.

\subsection{Intervention Layers}
We target middle layers 8-15 (inclusive), following evidence that middle layers are most critical for syntactic and semantic processing \citep{tenney2019bert, jawahar2019does}. Perturbations are applied to all token positions simultaneously at each targeted layer.

\subsection{Perturbation Magnitudes}
We test displacement values $\delta \in \{1.0, 2.0, 5.0, 10.0, 15.0, 20.0\}$, spanning from minimal perturbations to substantial interventions. Given that hidden state norms average approximately 15-25 in Pythia-410M, $\delta=5$ represents roughly a 20-30\% displacement relative to typical vector magnitudes.
\section{Results}
\label{sec:results}

\subsection{Loss Dissociation: Angular Perturbations Dominate}

Our first major finding is that angular perturbations cause dramatically more damage to language modeling loss than magnitude perturbations of equal size.

\begin{table}[t]
\centering
\small
\begin{tabular}{@{}ccccc@{}}
\toprule
$\delta$ & Mag.\ $\Delta$ & Ang.\ $\Delta$ & Ratio & $p$ \\
\midrule
1.0 & 0.009 {\scriptsize$\pm$0.002} & 0.368 {\scriptsize$\pm$0.011} & \textbf{42.9$\times$} & $<$0.001 \\
2.0 & 0.042 {\scriptsize$\pm$0.002} & 0.983 {\scriptsize$\pm$0.026} & \textbf{23.2$\times$} & $<$0.001 \\
5.0 & 0.700 {\scriptsize$\pm$0.028} & 3.757 {\scriptsize$\pm$0.060} & \textbf{5.4$\times$} & $<$0.001 \\
10.0 & 3.262 {\scriptsize$\pm$0.033} & 7.061 {\scriptsize$\pm$0.044} & \textbf{2.2$\times$} & $<$0.001 \\
15.0 & 4.600 {\scriptsize$\pm$0.045} & 7.718 {\scriptsize$\pm$0.050} & \textbf{1.7$\times$} & $<$0.001 \\
20.0 & 5.433 {\scriptsize$\pm$0.033} & 7.750 {\scriptsize$\pm$0.087} & \textbf{1.4$\times$} & $<$0.001 \\
\bottomrule
\end{tabular}
\caption{\textbf{Loss damage by perturbation type.} Angular perturbations cause dramatically more damage than magnitude perturbations at equal displacement $\delta$. Baseline loss: 4.107. All differences significant at $p < 0.001$ with large effect sizes.}
\label{tab:loss}
\end{table}

Table~\ref{tab:loss} presents the complete results. At the smallest perturbation ($\delta = 1.0$), angular perturbations cause \textbf{42.9 times} more loss increase than magnitude perturbations (0.368 vs.\ 0.009). This ratio decreases as $\delta$ increases, but remains substantial (2.2$\times$) even at $\delta = 10$.

\textbf{Two-Regime Analysis.} We observe two distinct regimes:
\begin{itemize}
    \item \textbf{Low-$\delta$ regime} ($\delta \leq 5$): Mean ratio 6.80$\times$, indicating that direction is disproportionately important for small displacements.
    \item \textbf{High-$\delta$ regime} ($\delta \geq 10$): Mean ratio 1.69$\times$, as both perturbations approach a ceiling where the model is maximally disrupted.
\end{itemize}

The 75\% reduction in ratio from low to high $\delta$ suggests that angular effects saturate: there is a limit to how much damage direction corruption can cause, likely because the model cannot produce worse than random predictions.

\textbf{Interpretation.} These results strongly suggest that direction encodes critical information for next token prediction. Rotating a vector even slightly fundamentally changes the information it conveys to downstream computation, leading to cascading errors. In contrast, scaling a vector preserves its informational content while modulating its influence.

\subsection{Syntactic Dissociation: Magnitude Perturbations Dominate}

Remarkably, the pattern \textit{reverses} for syntactic processing. Magnitude perturbations cause substantially more damage to subject-verb agreement accuracy than angular perturbations.

\begin{table}[t]
\centering
\small
\begin{tabular}{@{}cccccc@{}}
\toprule
$\delta$ & Mag.\ Acc. & Ang.\ Acc. & Mag.\ Drop & Ang.\ Drop & $p$ \\
\midrule
5.0 & 69.1\% {\scriptsize$\pm$2.2} & 87.9\% {\scriptsize$\pm$1.6} & \textbf{20.4\%} & 1.6\% & $<$0.0001 \\
10.0 & 56.0\% {\scriptsize$\pm$5.3} & 77.1\% {\scriptsize$\pm$2.0} & \textbf{33.5\%} & 12.4\% & 0.002 \\
15.0 & 53.5\% {\scriptsize$\pm$2.3} & 67.4\% {\scriptsize$\pm$2.6} & \textbf{36.0\%} & 22.1\% & $<$0.0001 \\
\bottomrule
\end{tabular}
\caption{\textbf{BLiMP syntactic accuracy.} Magnitude perturbations cause dramatically larger accuracy drops than angular perturbations. Baseline: 89.5\%. This pattern is opposite to loss damage (Table~\ref{tab:loss}).}
\label{tab:blimp}
\end{table}

Table~\ref{tab:blimp} shows that at $\delta = 5$, magnitude perturbations drop accuracy by 20.4\% while angular perturbations drop it by only 1.6\%---a \textbf{12.8$\times$ difference} in the opposite direction from loss effects. At $\delta = 10$, the pattern persists: 33.5\% vs.\ 12.4\% drop.

\subsection{The Double Dissociation}

\begin{figure}[t]
\centering
\includegraphics[width=\columnwidth]{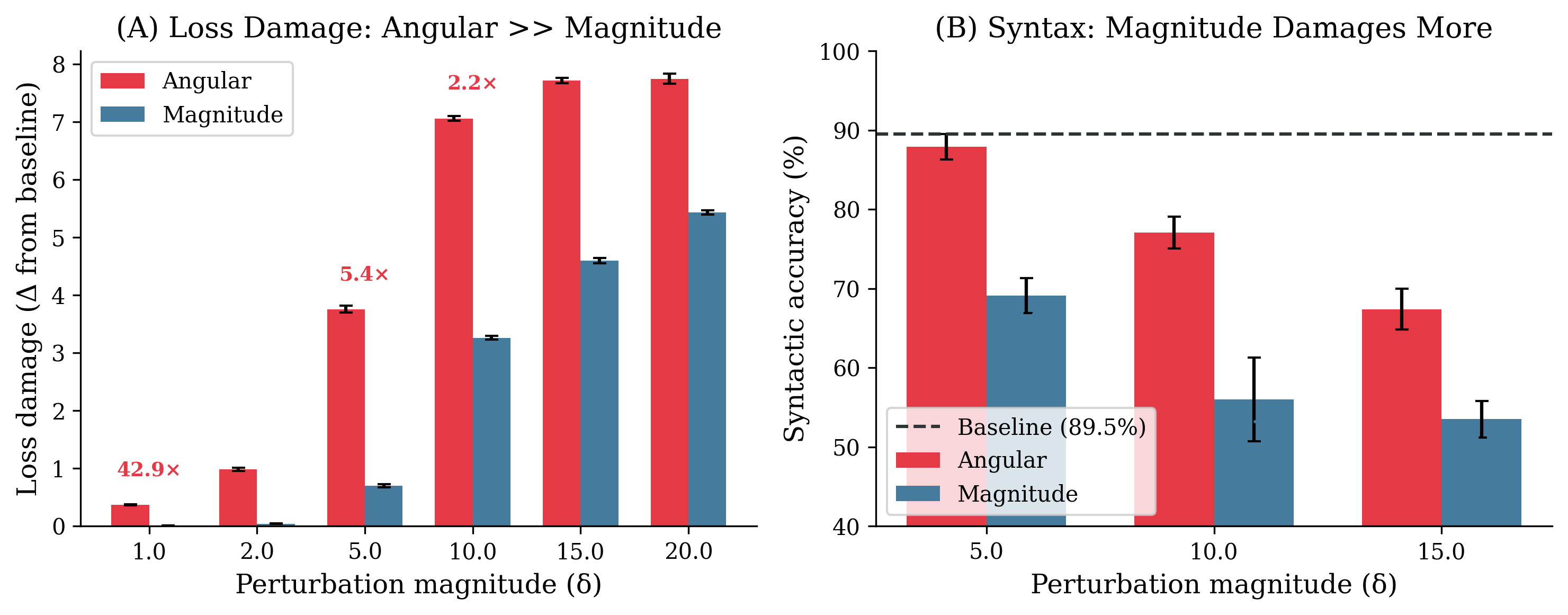}
\caption{\textbf{Cross-over dissociation.} (A) Loss damage across $\delta$ values: angular perturbations (red) cause up to 42.9$\times$ more damage than magnitude perturbations (blue) at small displacements. (B) BLiMP accuracy drop: the pattern reverses, with magnitude perturbations causing 12.8$\times$ greater accuracy loss at $\delta=5$. Error bars show standard error across 5 seeds.}
\label{fig:crossover}
\end{figure}

Tables~\ref{tab:loss} and \ref{tab:blimp} together constitute a \textbf{cross-over dissociation}, visualized in Figure~\ref{fig:crossover}:

\begin{table}[h]
\centering
\small
\begin{tabular}{@{}lcc@{}}
\toprule
Perturbation & Loss Damage & Syntax Damage \\
\midrule
\textbf{Angular} & \textbf{HIGH} (42$\times$) & LOW \\
\textbf{Magnitude} & LOW & \textbf{HIGH} (12$\times$) \\
\bottomrule
\end{tabular}
\caption{\textbf{Cross-over dissociation summary.} Each perturbation type preferentially impairs a different function, providing evidence for partial functional separation.}
\label{tab:double}
\end{table}

In cognitive neuroscience, such cross-over dissociations provide evidence for functional modularity \citep{shallice1988neuropsychology}. While both perturbation types cause some damage to both metrics at larger $\delta$, the \textit{preferential} pattern ,where each perturbation type disproportionately impairs a different function suggests that direction and magnitude support partially distinct computational roles in LayerNorm-based language models. As we discuss in Section~\ref{sec:limitations}, this pattern does not generalize to all architectures.

\subsection{Propagation Through Layers}

Although perturbations are L2-matched at the intervention layers, they propagate differently through subsequent computation. This differential propagation is itself informative about how the network processes direction versus magnitude.

\begin{table}[t]
\centering
\small
\begin{tabular}{@{}lccc@{}}
\toprule
Layer & Angular L2 & Magnitude L2 & Ratio \\
\midrule
8 (intervention start) & 5.00 & 5.00 & 1.00$\times$ \\
15 (intervention end) & 35.9 & 12.7 & 2.82$\times$ \\
23 (final) & 123.8 & 38.9 & 3.18$\times$ \\
\bottomrule
\end{tabular}
\caption{\textbf{Propagation amplification at $\delta=5$.} Angular perturbations amplify 3.18$\times$ more than magnitude perturbations by the final layer, despite identical L2 displacement at intervention.}
\label{tab:propagation}
\end{table}

Table~\ref{tab:propagation} shows that angular perturbations amplify 24.8$\times$ from intervention to final layer ($5.0 \rightarrow 123.8$), while magnitude perturbations amplify only 7.8$\times$ ($5.0 \rightarrow 38.9$). By the final layer, angular displacements are 3.18$\times$ larger than magnitude displacements despite starting at identical L2 distances.

We acknowledge that this differential propagation means our perturbations are no longer L2-matched at downstream layers, which partially contributes to the observed loss dissociation. However, we also interpret the differential propagation itself as evidence that the network treats directional errors differently from magnitude errors. Directional changes cascade through attention computation (which depends on cosine similarity between queries and keys), while magnitude changes are partially absorbed by LayerNorm (which renormalizes vector lengths). We provide causal evidence for these pathway differences in Section~\ref{sec:mechanistic}.

\subsection{Scale Generalization}

To assess whether the dissociation pattern generalizes beyond Pythia-410M, we replicate key experiments on Pythia-1.4B (3.5$\times$ larger).

\begin{table}[t]
\centering
\small
\begin{tabular}{@{}ccccc@{}}
\toprule
$\delta$ & 410M Ratio & 1.4B Ratio & $p$ (1.4B) \\
\midrule
2.0 & 23.2$\times$ & \textbf{56.8$\times$} & 0.0003 \\
5.0 & 5.4$\times$ & \textbf{24.4$\times$} & 0.0001 \\
10.0 & 2.2$\times$ & \textbf{6.0$\times$} & 0.0001 \\
\bottomrule
\end{tabular}
\caption{\textbf{Scale replication.} The angular/magnitude loss ratio is 2 to 4$\times$ higher in Pythia-1.4B than Pythia-410M, suggesting direction becomes relatively more important at scale.}
\label{tab:scale}
\end{table}

\begin{figure}[t]
\centering
\includegraphics[width=\columnwidth]{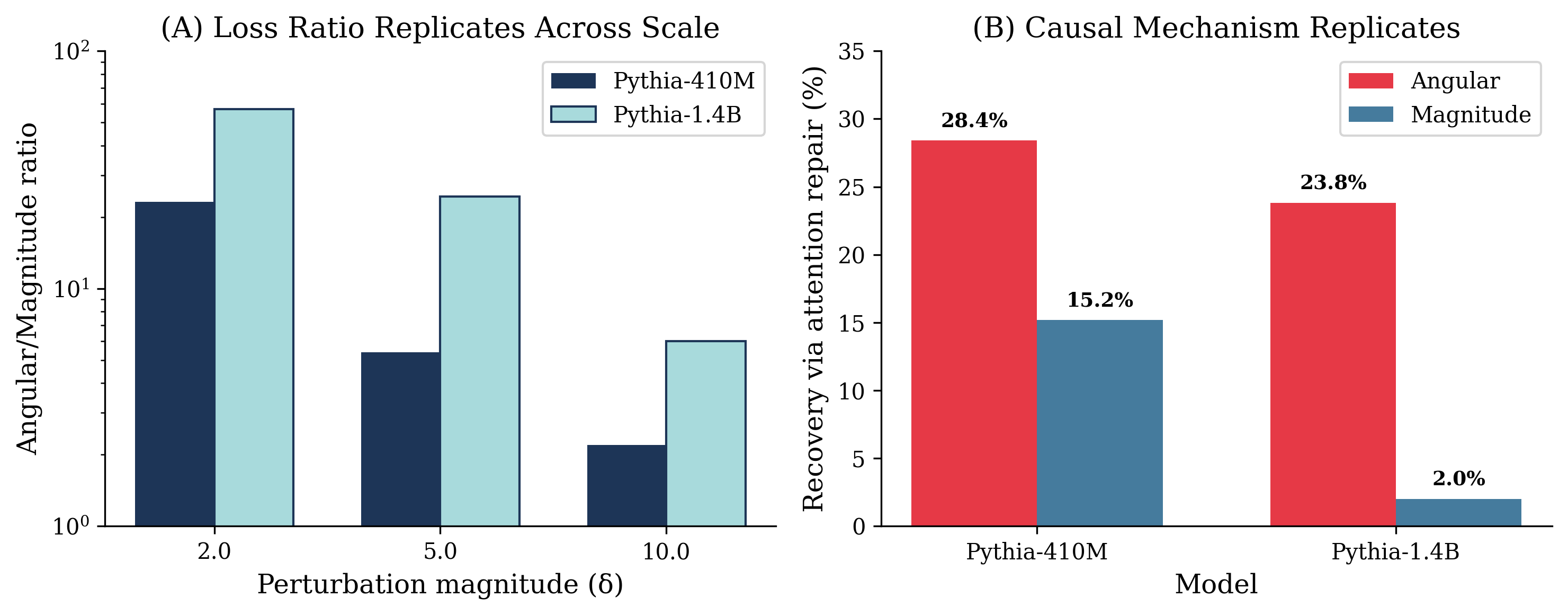}
\caption{\textbf{Dissociation replicates and amplifies at scale.} (A) Angular/magnitude loss damage ratio (log scale) across perturbation magnitudes. The effect is 2 to 4$\times$ stronger in Pythia-1.4B (light blue) than Pythia-410M (dark blue). (B) Causal attention repair pattern replicates across scales: angular recovery consistently exceeds magnitude recovery (28.4\% vs 15.2\% at 410M; 23.8\% vs 2.0\% at 1.4B).}
\label{fig:scale}
\end{figure}

Table~\ref{tab:scale} and Figure~\ref{fig:scale} show that the loss dissociation not only replicates but \textit{amplifies} at larger scale: the angular/magnitude ratio at $\delta=2$ increases from 23.2$\times$ (410M) to 56.8$\times$ (1.4B). This suggests that larger models may rely even more heavily on directional information for language modeling, though we caution that this observation is based on only two model scales within the same architecture family.

The mechanistic pattern also replicates qualitatively: causal attention repair recovers 23.8\% of angular damage in Pythia-1.4B (compared to 28.4\% in 410M), while magnitude damage recovery via attention repair drops to only 2.0\% (from 15.2\% in 410M). This consistency across scales within the Pythia family strengthens our confidence that the dissociation reflects genuine functional separation within this architecture. However, as detailed in Section~\ref{sec:limitations}, the pattern differs in other architecture families.

\section{Mechanistic Analysis}
\label{sec:mechanistic}

Having established the double dissociation, we now investigate the \textit{mechanisms} by which direction and magnitude affect downstream computation. Our central hypothesis is that direction operates substantially through attention, while magnitude operates through LayerNorm-mediated pathways.

\subsection{Attention Entropy}

We first examine whether perturbations affect the attention mechanism differently. We compute the Shannon entropy of attention distributions:
\begin{align}
    H(\mathbf{a}) = -\sum_i a_i \log a_i
\end{align}
Higher entropy indicates more diffuse attention; lower entropy indicates more focused attention.

\begin{table}[t]
\centering
\small
\begin{tabular}{@{}lcc@{}}
\toprule
Condition & Entropy & $\Delta$ from Baseline \\
\midrule
Baseline & 1.031 {\scriptsize$\pm$0.000} & — \\
Angular & 1.160 {\scriptsize$\pm$0.008} & \textbf{+0.129} {\scriptsize$\pm$0.008} \\
Magnitude & 1.056 {\scriptsize$\pm$0.001} & +0.025 {\scriptsize$\pm$0.001} \\
\bottomrule
\end{tabular}
\caption{\textbf{Attention entropy at $\delta=5$ (Pythia-410M).} Angular perturbations increase entropy 5.2$\times$ more than magnitude perturbations ($t=25.4$, $p<0.001$). Note: this pattern does not replicate at 1.4B scale; see Section~\ref{sec:causal_repair}.}
\label{tab:entropy}
\end{table}

Table~\ref{tab:entropy} shows that in Pythia-410M, angular perturbations increase attention entropy by \textbf{5.2 times} more than magnitude perturbations (0.129 vs.\ 0.025). This is consistent with direction being important for computing attention: rotating hidden states disrupts which tokens attend to which.

\textbf{Caveat.} While entropy provides suggestive correlational evidence, we note that this specific metric does not generalize reliably across scales—in Pythia-1.4B, the entropy pattern differs (angular perturbations slightly \textit{decrease} entropy). We therefore prioritize causal intervention evidence below.

\subsection{Causal Attention Repair}
\label{sec:causal_repair}

Correlational evidence does not establish that attention disruption \textit{causes} the observed loss damage. We therefore conduct a causal intervention: if angular damage operates through attention, then \textit{repairing} attention should recover lost performance.

\textbf{Design.} We perform two forward passes:
\begin{enumerate}
    \item \textbf{Clean pass:} No perturbation; cache all attention outputs
    \item \textbf{Repair pass:} Apply perturbation, but inject cached (clean) attention outputs
\end{enumerate}
If attention mediates the effect, the repair pass should show reduced damage.

\begin{table}[t]
\centering
\small
\begin{tabular}{@{}lccc@{}}
\toprule
Condition & Loss & Damage & Recovery \\
\midrule
Baseline & 4.107 & — & — \\
\midrule
Angular (unrepaired) & 7.715 {\scriptsize$\pm$0.088} & +3.608 & — \\
Angular (repaired) & 6.690 {\scriptsize$\pm$0.046} & +2.583 & \textbf{28.4\%} \\
\midrule
Magnitude (unrepaired) & 4.787 {\scriptsize$\pm$0.038} & +0.680 & — \\
Magnitude (repaired) & 4.683 {\scriptsize$\pm$0.035} & +0.576 & 15.2\% \\
\bottomrule
\end{tabular}
\caption{\textbf{Causal attention repair (Pythia-410M).} Repairing attention recovers significantly more loss for angular perturbations (28.4\%) than magnitude perturbations (15.2\%), $t=4.93$, $p=0.004$.}
\label{tab:repair}
\end{table}

Table~\ref{tab:repair} presents the causal repair results. Repairing attention recovers \textbf{28.4\%} of angular-induced loss but only \textbf{15.2\%} of magnitude-induced loss. This difference is statistically significant ($t = 4.93$, $p = 0.004$) and provides evidence that angular damage flows substantially through attention pathways. We note that attention repair is an imperfect intervention injecting clean attention outputs also affects downstream MLP inputs, so we interpret this as evidence that attention is \textit{a major pathway} rather than \textit{the sole pathway} for angular effects.

\textbf{Scale Replication.} Unlike the entropy metric, this causal pattern replicates at 1.4B scale: attention repair recovers 23.8\% of angular damage versus only 2.0\% for magnitude damage. The consistency of causal repair results across scales, despite inconsistent entropy patterns, underscores the importance of interventional over correlational evidence.

\textbf{Layer Specificity.} We further analyzed which layers contribute most to recovery. Repairing attention in later layers (12--15) recovers 25.0\% of damage, while early layers (8-11) account for only 6.4\% recovery ($t = 40.5$, $p < 0.0001$). This suggests that directional information becomes increasingly important for attention computation in later layers.
\begin{figure}[t]
\centering
\includegraphics[width=\columnwidth]{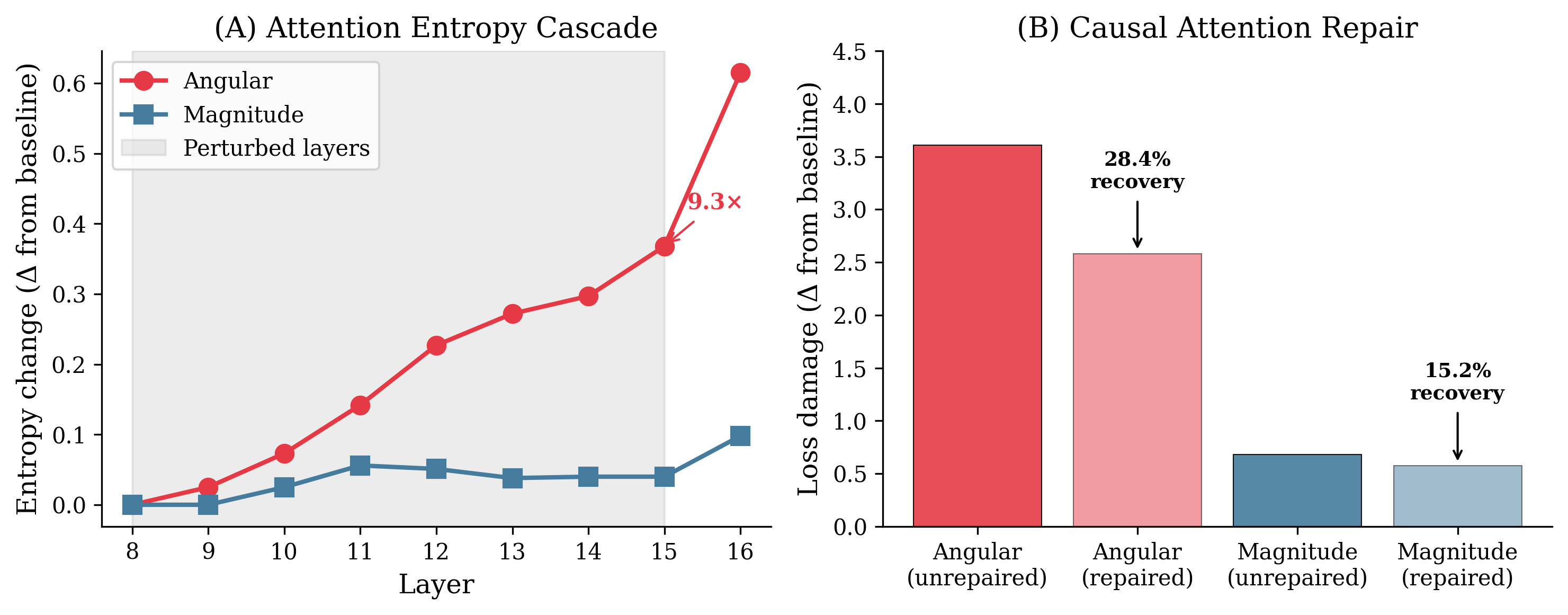}
\caption{\textbf{Attention mediates angular damage.} (A) Attention entropy change cascades through layers 8-16, with angular perturbations (red) causing 9.3$\times$ greater entropy increase than magnitude perturbations (blue) by layer 15. Shaded region indicates perturbed layers. (B) Causal attention repair recovers 28.4\% of angular-induced loss damage versus only 15.2\% for magnitude ($p=0.004$), confirming attention substantially mediates angular effects.}
\label{fig:attention}
\end{figure}

\subsection{Causal LayerNorm Repair}

If magnitude damage operates through attention-independent pathways, what is that pathway? We hypothesize that LayerNorm which explicitly manipulates vector norms may mediate magnitude effects. We test this with an analogous repair experiment.

\textbf{Design.} We cache clean LayerNorm outputs (both input LayerNorm and post-attention LayerNorm) for layers 8-15, then inject these cached outputs during perturbed forward passes.

\begin{table}[t]
\centering
\small
\begin{tabular}{@{}lccc@{}}
\toprule
Condition & Loss & Damage & Recovery \\
\midrule
Angular (unrepaired) & 7.816 & +3.605 & — \\
Angular (LN repaired) & 7.322 & +3.112 & 13.7\% \\
\midrule
Magnitude (unrepaired) & 4.890 & +0.680 & — \\
Magnitude (LN repaired) & 4.686 & +0.476 & \textbf{29.9\%} \\
\bottomrule
\end{tabular}
\caption{\textbf{Causal LayerNorm repair.} LayerNorm repair recovers significantly more damage for magnitude perturbations (29.9\%) than angular perturbations (13.7\%), $t=6.4$, $p=0.002$ the inverse of attention repair.}
\label{tab:ln_repair}
\end{table}

\begin{figure}[t]
\centering
\includegraphics[width=\columnwidth]{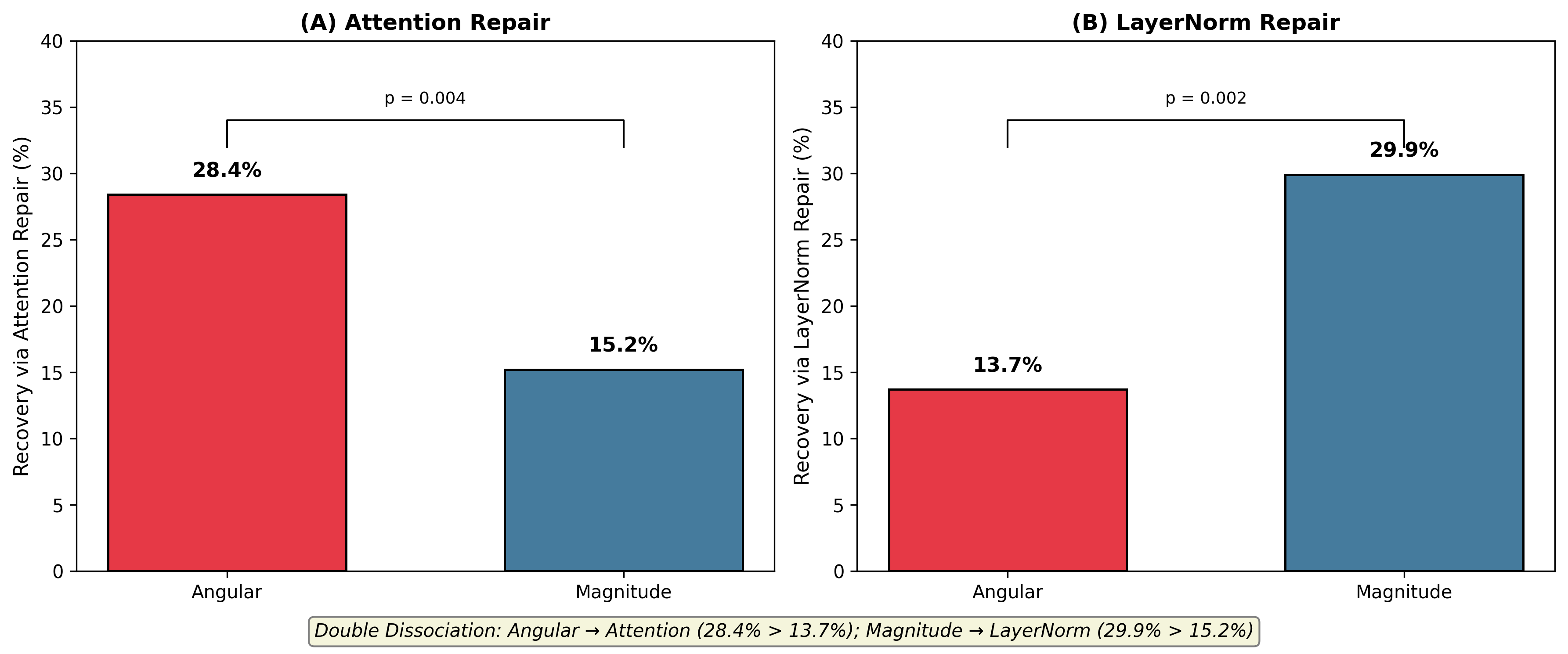}
\caption{\textbf{Double dissociation in mechanistic pathways.} (A) Attention repair preferentially recovers angular damage (28.4\%) over magnitude damage (15.2\%), $p=0.004$. (B) LayerNorm repair shows the inverse pattern: magnitude recovery (29.9\%) exceeds angular recovery (13.7\%), $p=0.002$. This mechanistic double dissociation parallels the behavioral dissociation in Figure~\ref{fig:crossover}.}
\label{fig:pathway_dissociation}
\end{figure}

Table~\ref{tab:ln_repair} and Figure~\ref{fig:pathway_dissociation} show that LayerNorm repair recovers \textbf{29.9\%} of magnitude induced damage but only \textbf{13.7\%} of angular induced damage ($t = 6.4$, $p = 0.002$). This is the \textit{inverse} of the attention repair pattern, providing causal evidence for distinct mechanistic pathways:

\begin{itemize}
    \item \textbf{Angular damage}: Substantially mediated by attention (28.4\% recovery) but not LayerNorm (13.7\%)
    \item \textbf{Magnitude damage}: Substantially mediated by LayerNorm (29.9\% recovery) but not attention (15.2\%)
\end{itemize}

\subsection{Why Does Magnitude Affect Syntax?}

A puzzle remains: why do magnitude perturbations specifically impair syntactic processing? We investigate by examining what magnitude encodes.

\textbf{Parse Depth Correlation.} Following \citet{hewitt2019structural}, who showed that syntax trees can be recovered from representation geometry, we test whether vector magnitude correlates with parse tree depth (distance from root in the dependency tree).

\begin{table}[t]
\centering
\small
\begin{tabular}{@{}lcc@{}}
\toprule
Condition & Pearson $r$ & Change \\
\midrule
Baseline & +0.142 ($p < 10^{-45}$) & — \\
Angular perturbation & +0.126 & $-$11\% \\
Magnitude perturbation & +0.036 & $-$75\% \\
\bottomrule
\end{tabular}
\caption{\textbf{Magnitude correlates with parse depth.} Baseline representations show significant positive correlation between $\|\mathbf{h}\|$ and parse tree depth. Magnitude perturbation destroys this correlation (75\% reduction), while angular perturbation largely preserves it.}
\label{tab:parse_depth}
\end{table}

\begin{figure}[t]
\centering
\includegraphics[width=\columnwidth]{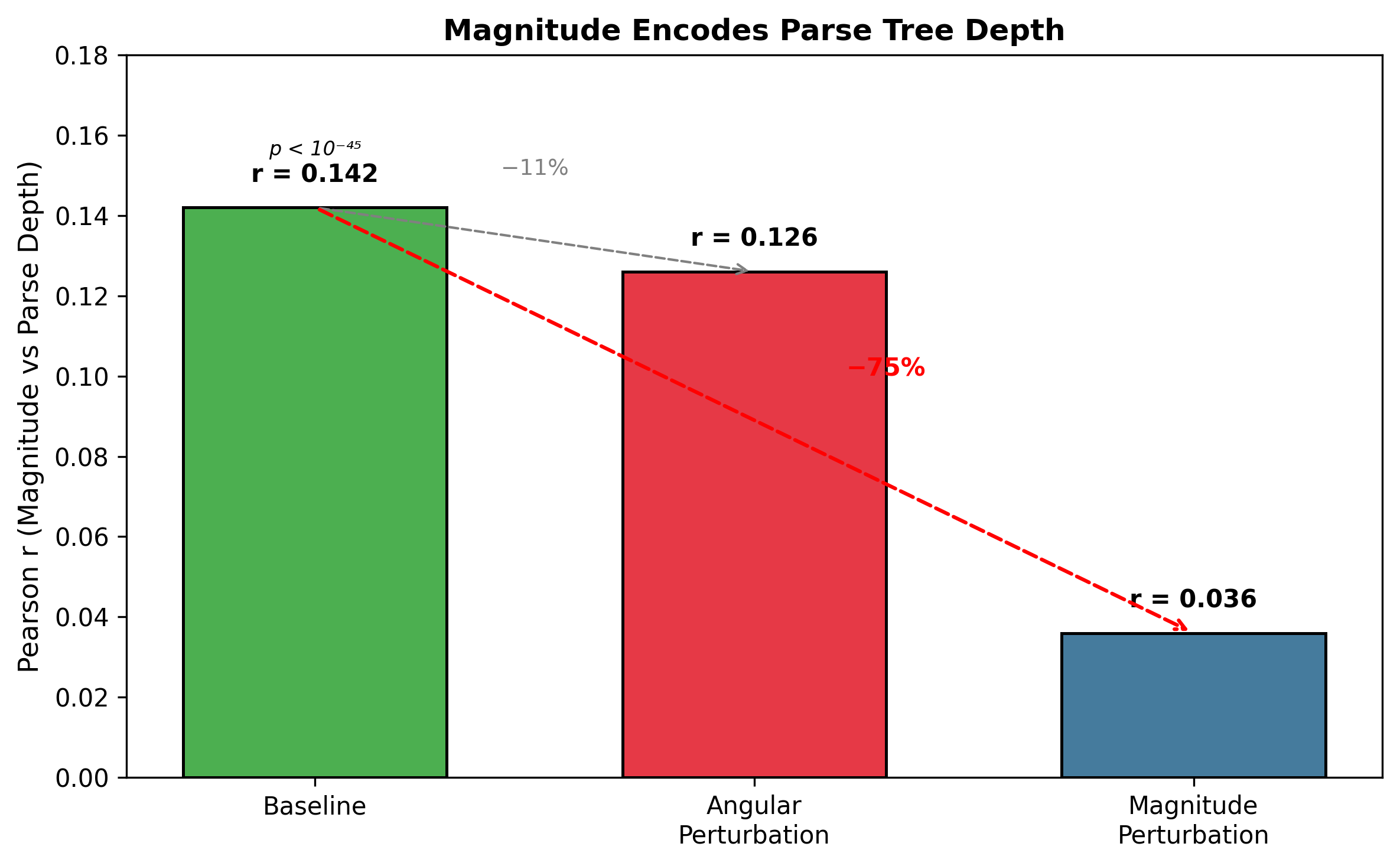}
\caption{\textbf{Magnitude encodes parse tree depth.} Pearson correlation between vector magnitude $\|\mathbf{h}\|$ and dependency parse depth. Baseline shows significant positive correlation ($r=0.142$, $p<10^{-45}$). Angular perturbation preserves this structure ($-11\%$), while magnitude perturbation destroys it ($-75\%$), explaining why magnitude perturbations specifically impair syntactic computation that depends on structural position.}
\label{fig:parse_depth}
\end{figure}

Table~\ref{tab:parse_depth} and Figure~\ref{fig:parse_depth} show that magnitude correlates significantly with parse tree depth ($r = 0.142$, $p < 10^{-45}$): tokens deeper in the syntactic tree have larger magnitude. Critically, magnitude perturbation reduces this correlation by 75\% (to $r = 0.036$), while angular perturbation reduces it by only 11\%.

This provides a mechanistic explanation for why magnitude perturbations impair syntax: magnitude encodes structural position information needed for syntactic computation. Disrupting magnitude destroys the model's representation of where tokens sit in the parse tree, which is critical for subject-verb agreement (the model must identify which noun is the subject).
\subsection{Encoding vs Computation: Resolving an Apparent Paradox}

Linear probing reveals an apparent paradox: direction encodes more syntactic information, yet magnitude perturbations damage syntax more.

\begin{table}[t]
\centering
\small
\begin{tabular}{@{}lcc@{}}
\toprule
Representation & POS Accuracy & vs.\ Chance (16.7\%) \\
\midrule
Full vector $\mathbf{h}$ & 82.3\% & +65.6\% \\
Direction only ($\hat{\mathbf{h}}$) & 81.6\% {\scriptsize$\pm$0.5} & +64.9\% \\
Magnitude only ($\|\mathbf{h}\|$) & 29.7\% {\scriptsize$\pm$4.0} & +13.0\% \\
\bottomrule
\end{tabular}
\caption{\textbf{POS probing.} Direction carries 2.7$\times$ more syntactic information than magnitude (81.6\% vs.\ 29.7\%), yet magnitude perturbations damage syntax more (Table~\ref{tab:blimp}).}
\label{tab:probing}
\end{table}

Table~\ref{tab:probing} shows that direction alone achieves 81.6\% POS accuracy (only 0.7\% below full vector), while magnitude alone achieves 29.7\%-2.7$\times$ less syntactic information.

How do we reconcile this with Table~\ref{tab:blimp}, which shows magnitude perturbations damage syntax more? The resolution lies in distinguishing \textit{encoding} from \textit{computation}:

\begin{itemize}
    \item \textbf{Direction encodes} syntactic categories (what POS a token is)
    \item \textbf{Magnitude modulates} how that information is used in computation (via parse depth encoding and LayerNorm mediated processing)
\end{itemize}

Supporting this interpretation, we find that the direction probe is robust to perturbation: after angular perturbation, the direction probe still achieves 79.9\% accuracy ($-$1.7\% drop), indicating the syntactic encoding survives. The damage to syntax comes not from destroying the encoding but from disrupting how the model \textit{uses} that encoding and magnitude perturbations specifically disrupt the structural position information (parse depth) needed to compute agreement.

\subsection{Summary: Two Mechanistic Pathways}

\begin{figure}[t]
\centering
\includegraphics[width=\columnwidth]{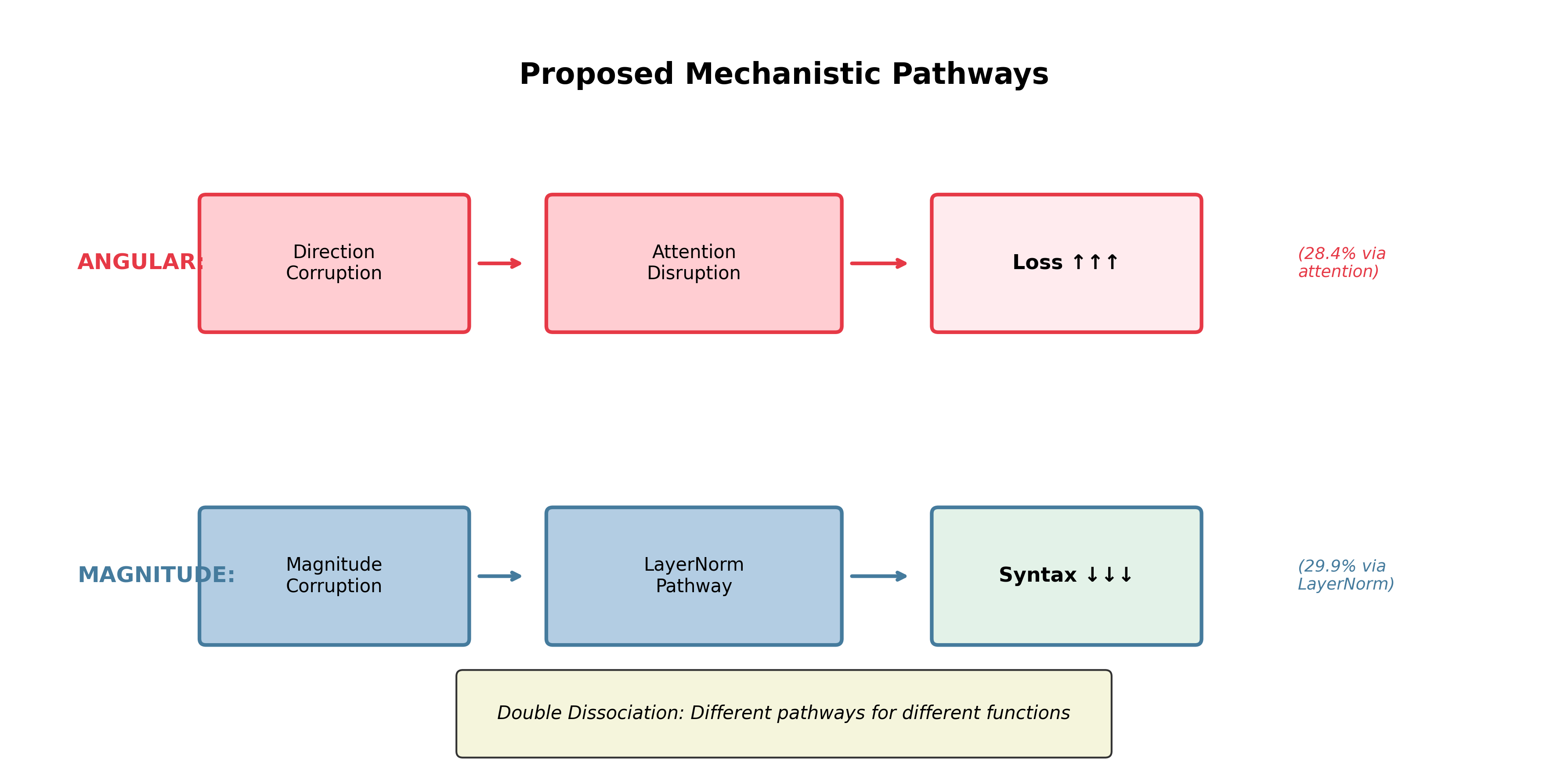}
\caption{\textbf{Proposed mechanistic pathways.} Angular perturbations (top, red) corrupt direction, disrupt attention patterns, and cause cascading loss damage. Magnitude perturbations (bottom, blue) corrupt magnitude, disrupt LayerNorm-mediated processing, and specifically impair syntactic computation. The double dissociation in pathways (attention vs LayerNorm) mirrors the behavioral double dissociation (loss vs syntax).}
\label{fig:pathways_diagram}
\end{figure}

Our findings support a model with two partially distinct mechanistic pathways (Figure~\ref{fig:pathways_diagram}):

\begin{table}[h]
\centering
\small
\begin{tabular}{@{}lcc@{}}
\toprule
Pathway & Attention Recovery & LayerNorm Recovery \\
\midrule
Angular & \textbf{28.4\%} & 13.7\% \\
Magnitude & 15.2\% & \textbf{29.9\%} \\
\bottomrule
\end{tabular}
\caption{\textbf{Double dissociation in mechanistic pathways.} Angular damage is preferentially recovered by attention repair; magnitude damage is preferentially recovered by LayerNorm repair.}
\label{tab:pathway_summary}
\end{table}

\textbf{Direction $\rightarrow$ Attention pathway:} Direction substantially influences attention patterns (which tokens attend to which). Corrupting direction disrupts information routing, causing cascading errors that propagate through layers and damage language modeling.

\textbf{Magnitude $\rightarrow$ LayerNorm pathway:} Magnitude correlates with structural position (parse depth) and appears to modulate processing intensity via LayerNorm. Corrupting magnitude disrupts structural encoding needed for fine-grained syntactic computation like subject-verb agreement.

We emphasize that these pathways account for roughly 30\% of each perturbation type's damage. The remaining $\sim$70\% likely flows through MLP layers and residual connections, which we did not repair. The complete mechanistic picture remains an important direction for future work.
\section{Generalization}

\subsection{Across Layers}
We verified that direction carries more syntactic information than magnitude across all layers tested (0, 4, 8, 12, 16, 20, 23), with direction achieving 75-83\% POS accuracy versus magnitude's 19-40\% (see Appendix).

\subsection{Architecture Generalization: A Critical Caveat}
We tested generalization to other architectures with mixed results that significantly qualify our claims:

\begin{itemize}
    \item \textbf{OPT-1.3B (LayerNorm):} The loss dissociation replicates (32.7$\times$ at $\delta=2$), but the attention mediation pattern \textit{reverses} magnitude perturbations show higher attention recovery (34.4\%) than angular (27.0\%).
    \item \textbf{TinyLlama-1.1B (RMSNorm):} The loss dissociation \textit{reverses entirely} magnitude perturbations cause approximately 5$\times$ more damage than angular at low $\delta$, the opposite of our Pythia findings.
\end{itemize}

These results suggest that our findings are specific to the Pythia architecture family and may depend on the choice of normalization layer. The direction-magnitude dissociation should be interpreted as a property of LayerNorm-based transformers in the GPT-NeoX family, not as a universal property of all transformer architectures. We consider this architecture dependence to be a significant finding in itself, suggesting that normalization type may fundamentally affect how transformers utilize geometric properties of their representations.
\section{Discussion}

\subsection{Refining the Linear Representation Hypothesis}

Our findings suggest a refinement of the linear representation hypothesis:

\textbf{Extended Hypothesis (for LayerNorm-based architectures):} In transformers using standard LayerNorm, direction and magnitude serve \textit{partially distinct computational roles} direction predominantly governs attentional routing, while magnitude modulates processing intensity through LayerNorm-mediated pathways. This hypothesis may not apply to architectures using alternative normalization schemes such as RMSNorm.

Direction implements the core insight of the original hypothesis: concepts correspond to orientations in activation space. But magnitude is not merely noise or redundancy. Our evidence suggests that magnitude encodes structural position information (parse tree depth) and modulates how directional information influences downstream computation. We emphasize that this is a working hypothesis supported by our evidence, not a definitive claim.

\subsection{Encoding vs Computation: A Key Distinction}

Our most theoretically significant finding may be the dissociation between \textit{encoding} and \textit{computation}. Linear probing shows that direction encodes syntactic categories (81.6\% POS accuracy vs.\ 29.7\% for magnitude). Yet magnitude perturbations damage syntactic \textit{processing} more than angular perturbations.

This apparent paradox resolves when we distinguish what information a representation \textit{contains} from how that information is \textit{used}:
\begin{itemize}
    \item Direction encodes \textbf{categorical information} (what syntactic category a token is)
    \item Magnitude encodes \textbf{structural position} (where in the parse tree) and modulates how categorical information influences computation
\end{itemize}

This distinction has methodological implications: probing accuracy may not predict behavioral importance. A component with lower probing accuracy (magnitude) can still be critical for task performance if it mediates how encoded information is processed.

\subsection{Two Mechanistic Pathways}

Our causal repair experiments reveal a double dissociation in mechanistic pathways:

\begin{itemize}
    \item \textbf{Attention repairs angular damage} (28.4\% recovery) but not magnitude damage (15.2\%)
    \item \textbf{LayerNorm repairs magnitude damage} (29.9\% recovery) but not angular damage (13.7\%)
\end{itemize}

This suggests that direction and magnitude flow through partially distinct computational pathways. Direction affects attention patterns which tokens attend to which creating cascading effects when corrupted (as evidenced by 3$\times$ greater propagation amplification). Magnitude affects LayerNorm-mediated processing, disrupting the structural position encoding needed for syntactic computation.

We note that these pathways account for roughly 30\% of each perturbation type's damage. The remaining effects likely flow through MLP layers and residual connections, which we did not repair. Fully characterizing these pathways remains an important direction for future work.

\subsection{Implications for Model Editing}

Our findings have practical implications for techniques that edit model knowledge \citep{meng2022locating, mitchell2022fast}. Current methods typically modify activation vectors without distinguishing direction from magnitude. Our results suggest more targeted approaches:

\begin{itemize}
    \item \textbf{Preserve direction} when the goal is maintaining language modeling quality and factual knowledge (which likely depend on attentional routing)
    \item \textbf{Preserve magnitude} when the goal is maintaining syntactic coherence and structural computation
    \item \textbf{Consider separate edits} for direction vs.\ magnitude when modifying specific capabilities
\end{itemize}

More generally, understanding \textit{which} geometric properties matter for \textit{which} capabilities could enable more surgical interventions with fewer unintended side effects.

\subsection{Implications for Interpretability}

Current interpretability techniques probing, activation patching, circuit analysis typically operate on full vectors. Our results suggest that separately analyzing direction and magnitude components could reveal distinct information streams:

\begin{itemize}
    \item \textbf{Direction probing} may reveal categorical and semantic content
    \item \textbf{Magnitude probing} may reveal structural position and processing intensity signals
    \item \textbf{Separate patching} of direction vs.\ magnitude could enable finer-grained causal analysis
\end{itemize}

The decomposition $\mathbf{h} = \|\mathbf{h}\| \cdot \hat{\mathbf{h}}$ is simple to compute and may be a useful addition to the interpretability toolkit.


\section{Limitations}
\label{sec:limitations}

We acknowledge several limitations:

\begin{enumerate}
    \item \textbf{L2 matching assumption:} We assume equal Euclidean distance makes perturbations comparable. However, neural networks operate in highly anisotropic spaces where some directions matter far more than others. While we verify L2 matching at intervention layers, angular perturbations amplify 3$\times$ more than magnitude perturbations by the final layer (Section~\ref{sec:results}). Future work could explore alternative matching criteria using loss sensitivity directions or Hessian analysis.
    
    \item \textbf{Random rotation directions:} Our angular perturbations use random orthogonal directions as controlled geometric probes. Results might differ for systematic rotations toward or away from known feature directions (e.g., syntax-relevant subspaces identified via probing). Whether the dissociation holds for structured perturbations remains an open question.
    
    \item \textbf{Architecture generalization (critical limitation):} Our primary findings are on the Pythia family, and generalization tests reveal significant architecture dependence:
    \begin{itemize}
        \item \textbf{OPT-1.3B (LayerNorm):} The loss dissociation replicates (32.7$\times$ at $\delta=2$), but the attention mediation pattern reverses magnitude shows \textit{higher} attention recovery (34.4\%) than angular (27.0\%), opposite to Pythia.
        \item \textbf{TinyLlama-1.1B (RMSNorm):} The loss dissociation \textit{reverses entirely} magnitude perturbations cause approximately 5$\times$ more damage than angular at low $\delta$, the opposite of our main findings.
        \item \textbf{OPT-IML-1.3B (instruction-tuned):} Shows the same pattern as OPT base, ruling out instruction tuning as the cause of TinyLlama's reversal.
    \end{itemize}
    This strongly suggests our findings are specific to the Pythia/GPT-NeoX architecture family. The TinyLlama reversal in particular indicates that the direction-magnitude dissociation is \textbf{not} a universal transformer property but depends on architectural choices, likely the normalization type (LayerNorm vs.\ RMSNorm). We consider this architecture dependence to be an important finding that warrants future investigation.
    
    \item \textbf{Layer scope:} We target middle layers (8--15) based on prior work suggesting these layers are most active for syntactic and semantic processing. Early layers (embedding) and late layers (output projection) may show different patterns. While we analyze layer-wise propagation, we did not test the main dissociation with different intervention layer ranges.
    
    \item \textbf{Syntactic scope:} We evaluate syntactic processing using BLiMP subject-verb agreement, a single syntactic phenomenon. Other phenomena long-distance dependencies, garden-path sentences, binding, island constraints may show different magnitude sensitivity patterns. Whether the magnitude$\rightarrow$syntax link generalizes beyond agreement remains to be tested.
    
    \item \textbf{Mechanistic incompleteness:} Our causal repair experiments identify two pathways: attention mediates $\sim$28\% of angular damage, and LayerNorm mediates $\sim$30\% of magnitude damage. However, the remaining $\sim$70\% of each perturbation type's damage is unaccounted for and likely flows through MLP layers and residual connections. Fully characterizing these pathways remains important future work.
    
    \item \textbf{Causal intervention artifacts:} Attention and LayerNorm repair are imperfect interventions. Injecting clean attention outputs may also affect what information flows through MLP layers; injecting clean LayerNorm outputs may disrupt expected input distributions. More precise interventions (e.g., repairing individual heads or specific LayerNorm components) could strengthen causal claims.
    
    \item \textbf{Correlational metrics:} Attention entropy, while suggestive at 410M scale, does not generalize reliably the pattern reverses in Pythia-1.4B. We prioritize causal repair evidence over correlational metrics, but this inconsistency highlights that surface-level metrics may not reflect underlying mechanisms.
\end{enumerate}


\section{Future Directions}

Our work opens several research directions:

\textbf{Feature-Specific Rotations.} Rather than random rotations, systematically rotating toward or away from known feature directions (e.g., sentiment, factuality) could reveal how specific features depend on direction vs magnitude.

\textbf{Training Dynamics.} How do the roles of direction and magnitude evolve during training? Do models initially rely on magnitude before developing directional structure?

\textbf{Magnitude as Confidence.} Our finding that magnitude encodes structural position connects to uncertainty quantification. Can magnitude-based uncertainty estimates improve calibration?

\textbf{Intervention Design.} Our causal repair technique could be extended to design more surgical model edits that preserve either direction or magnitude while modifying the other.

\textbf{Other Modalities.} Vision transformers also use vector representations. Do direction and magnitude play similar roles in visual processing?

\textbf{Alternative Normalization.} Our finding that the dissociation reverses in TinyLlama (which uses RMSNorm) suggests that normalization type may determine whether magnitude carries functional information. Systematic comparison across normalization schemes could clarify this.

\section{Conclusion}

We introduced \textbf{L2-matched perturbation analysis}, a methodology enabling controlled comparison of direction versus magnitude in transformer representations by ensuring both perturbation types achieve identical Euclidean displacements. Applying this framework to Pythia-family models, we discovered a \textbf{cross-over dissociation}: angular perturbations cause up to 42.9$\times$ more damage to language modeling loss, while magnitude perturbations cause up to 20 percentage points greater drop in syntactic accuracy.

Mechanistic analysis revealed two partially distinct pathways: angular damage is substantially mediated by attention (28.4\% causal recovery), while magnitude damage is substantially mediated by LayerNorm (29.9\% causal recovery). This double dissociation in both behavior and mechanism provides evidence that direction and magnitude serve \textbf{partially distinct computational roles} direction preferentially governs attentional routing, while magnitude encodes structural position and modulates processing intensity through LayerNorm-mediated pathways.

These findings refine the linear representation hypothesis for LayerNorm-based architectures by showing that the geometry of hidden states is not merely a container for information but actively shapes how that information flows through computation. The reversal of this pattern in RMSNorm-based architectures (TinyLlama) suggests that architectural choices fundamentally affect the functional roles of geometric properties.

Our results have implications for interpretability research (separately analyzing direction and magnitude may reveal distinct information streams), model editing (preserving direction vs.\ magnitude depending on the target capability), and theoretical understanding of neural language models.

We emphasize that our findings are specific to the Pythia/OPT architecture family and do \textbf{not} generalize to all transformers. TinyLlama (RMSNorm-based) shows the opposite pattern magnitude perturbations cause more loss damage than angular indicating that the direction-magnitude dissociation depends critically on architectural choices such as normalization type. This architecture dependence is itself an important finding for future investigation.

The geometry of representation is not just a mathematical curiosity it is a window into the computational structure of neural language models, though the specific roles of geometric properties depend on architectural choices that remain to be fully understood.
\section*{Impact Statement}

This paper presents work whose goal is to advance the field of Machine Learning, specifically in the area of neural network interpretability. Our findings on how transformers encode and process information through geometric properties of their representations contribute to the fundamental understanding of these models. 

The methodological contributions (L2-matched perturbation analysis) and mechanistic findings (direction$\rightarrow$attention, magnitude$\rightarrow$LayerNorm pathways) provide tools that could help researchers better understand and potentially improve the reliability and predictability of language models. Understanding which geometric properties matter for which capabilities may eventually enable more targeted model editing with fewer unintended side effects.

There are many potential societal consequences of our work, none which we feel must be specifically highlighted here.

\bibliography{references}
\bibliographystyle{icml2026}

\newpage
\appendix

\section{Statistical Details}
\label{app:stats}

All experiments use 5 random seeds. We report means $\pm$ standard errors. Statistical significance is assessed via two-tailed paired t-tests with Bonferroni correction for multiple comparisons. We acknowledge that 5 seeds provides limited statistical power; however, effect sizes are consistently large across all primary comparisons.

\section{Perturbation Verification}
\label{app:verification}

We verified that all perturbations achieve target displacement $\delta$ within tolerance 0.01:

\begin{table}[h]
\centering
\small
\begin{tabular}{@{}ccc@{}}
\toprule
Target $\delta$ & Angular Achieved & Magnitude Achieved \\
\midrule
1.0 & 1.000 $\pm$ 0.001 & 1.000 $\pm$ 0.000 \\
5.0 & 5.000 $\pm$ 0.002 & 5.000 $\pm$ 0.001 \\
10.0 & 10.000 $\pm$ 0.003 & 10.000 $\pm$ 0.001 \\
\bottomrule
\end{tabular}
\caption{Verification that L2-matching achieves target displacement at intervention layers.}
\label{tab:verification}
\end{table}

\section{Propagation Through Layers}
\label{app:propagation}

Table~\ref{tab:prop_detail} shows how L2 displacement propagates through all layers after intervention at layers 8--15 with $\delta=5$.

\begin{table}[h]
\centering
\small
\begin{tabular}{@{}lccc@{}}
\toprule
Layer & Angular L2 & Magnitude L2 & Ratio \\
\midrule
8 (intervention start) & 5.00 & 5.00 & 1.00$\times$ \\
9 & 8.15 & 5.78 & 1.41$\times$ \\
10 & 12.3 & 7.02 & 1.75$\times$ \\
11 & 17.8 & 8.45 & 2.11$\times$ \\
12 & 23.1 & 9.67 & 2.39$\times$ \\
13 & 27.9 & 10.8 & 2.58$\times$ \\
14 & 32.0 & 11.7 & 2.74$\times$ \\
15 (intervention end) & 35.9 & 12.7 & 2.82$\times$ \\
20 & 78.4 & 26.2 & 2.99$\times$ \\
23 (final) & 123.8 & 38.9 & 3.18$\times$ \\
\bottomrule
\end{tabular}
\caption{L2 displacement propagation at $\delta=5$. Angular perturbations amplify 3.18$\times$ more than magnitude perturbations by the final layer, despite identical displacement at intervention start.}
\label{tab:prop_detail}
\end{table}

\section{Layer-wise Entropy Changes}
\label{app:entropy}

Table~\ref{tab:entropy_layers} shows attention entropy changes by layer in Pythia-410M at $\delta=5$. Note: this pattern does not replicate in Pythia-1.4B (see Section~\ref{sec:mechanistic}).

\begin{table}[h]
\centering
\small
\begin{tabular}{@{}cccc@{}}
\toprule
Layer & Angular $\Delta$ & Magnitude $\Delta$ & Ratio \\
\midrule
8 & +0.000 & +0.000 & — \\
9 & +0.025 & +0.000 & — \\
10 & +0.073 & +0.025 & 2.9$\times$ \\
11 & +0.142 & +0.056 & 2.5$\times$ \\
12 & +0.227 & +0.051 & 4.4$\times$ \\
13 & +0.272 & +0.038 & 7.2$\times$ \\
14 & +0.297 & +0.040 & 7.4$\times$ \\
15 & +0.368 & +0.040 & 9.3$\times$ \\
16 & +0.615 & +0.098 & 6.3$\times$ \\
\bottomrule
\end{tabular}
\caption{Layer-wise attention entropy change at $\delta=5$ (Pythia-410M). The angular/magnitude ratio increases through perturbed layers (8--15), indicating cascading effects through attention.}
\label{tab:entropy_layers}
\end{table}

\section{Layer-wise Probing Results}
\label{app:probing}

\begin{figure}[h]
\centering
\includegraphics[width=\columnwidth]{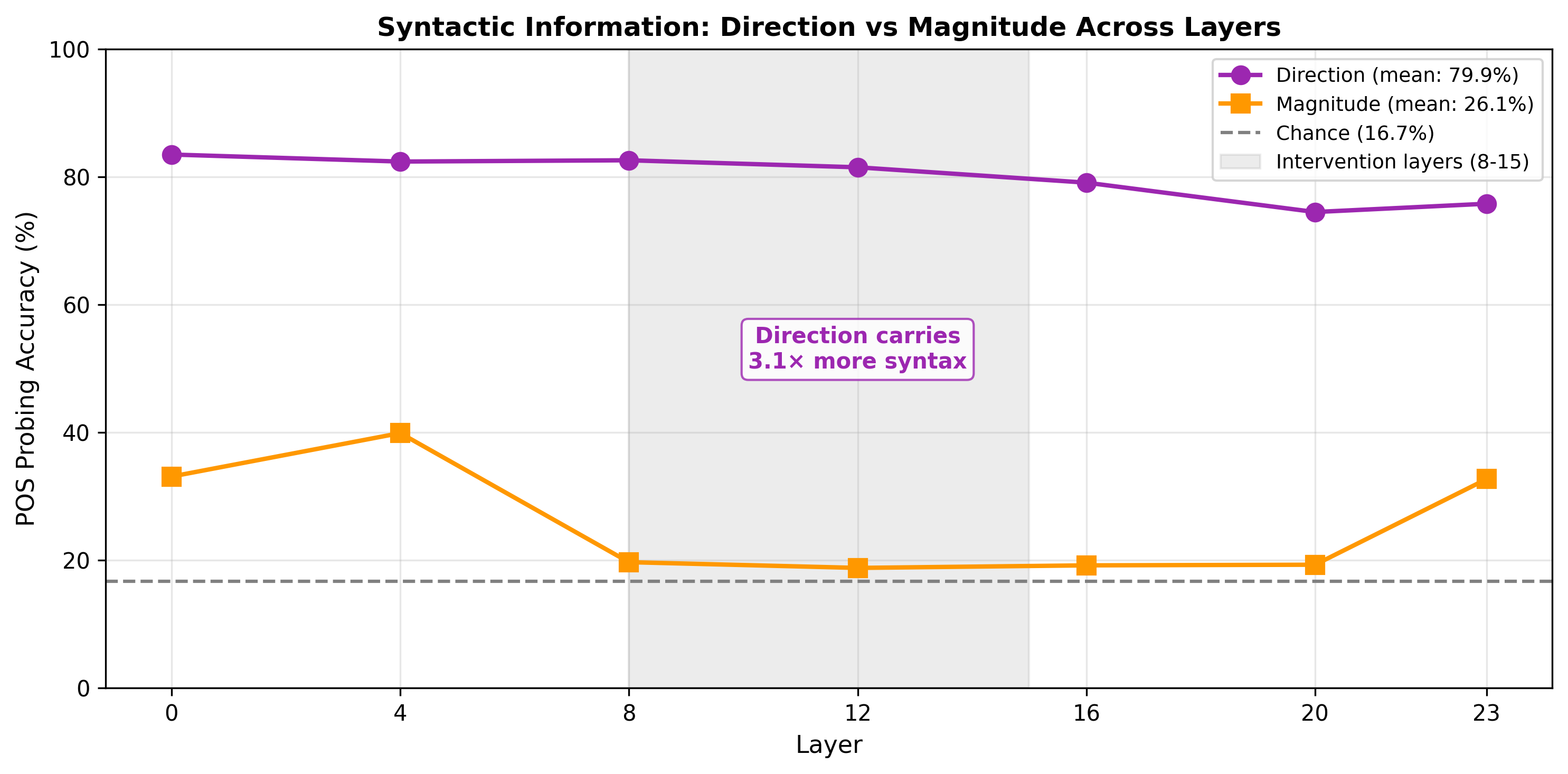}
\caption{\textbf{Direction encodes syntax across all layers.} POS probing accuracy for direction-only (purple) and magnitude-only (orange) representations across layers 0--23. Direction consistently outperforms magnitude (mean 79.9\% vs 26.1\%), with the gap largest in middle layers (8--16, shaded) where magnitude drops to near-chance (16.7\%, dashed line). This pattern holds across all tested layers.}
\label{fig:layerwise}
\end{figure}

Table~\ref{tab:probe_layers} and Figure~\ref{fig:layerwise} show POS probing accuracy across layers,confirming that direction carries more syntactic information than magnitude at all depths.

\begin{table}[h]
\centering
\small
\begin{tabular}{@{}lccc@{}}
\toprule
Layer & Full & Direction & Magnitude \\
\midrule
0 & 90.7\% & 83.5\% & 33.1\% \\
4 & 94.3\% & 82.4\% & 39.9\% \\
8 & 93.3\% & 82.6\% & 19.7\% \\
12 & 93.2\% & 81.5\% & 18.8\% \\
16 & 89.8\% & 79.1\% & 19.2\% \\
20 & 87.3\% & 74.5\% & 19.3\% \\
23 & 92.5\% & 75.8\% & 32.7\% \\
\midrule
Mean & 91.6\% & 79.9\% & 26.1\% \\
\bottomrule
\end{tabular}
\caption{POS probing accuracy by layer. Direction consistently outperforms magnitude (mean 79.9\% vs.\ 26.1\%), with magnitude near chance (16.7\%) in middle layers.}
\label{tab:probe_layers}
\end{table}
\section{Architecture Generalization Details}
\label{app:architecture}

Table~\ref{tab:arch} summarizes results across architectures tested.

\begin{table}[h]
\centering
\small
\begin{tabular}{@{}lccc@{}}
\toprule
Model & Loss Ratio ($\delta$=2) & Attention Recovery & Pattern \\
\midrule
Pythia-410M & 23.2$\times$ & Ang: 28.4\%, Mag: 15.2\% & $\checkmark$ Expected \\
Pythia-1.4B & 56.8$\times$ & Ang: 23.8\%, Mag: 2.0\% & $\checkmark$ Expected \\
OPT-1.3B & 32.7$\times$ & Ang: 27.0\%, Mag: 34.4\% & \textbf{(!)} Reversed mediation \\
TinyLlama-1.1B & 0.2$\times$ & --- & \textbf{(!)} Reversed dissociation \\
\bottomrule
\end{tabular}
\caption{Architecture generalization. The loss dissociation replicates in Pythia and OPT but reverses in TinyLlama. Attention mediation replicates in Pythia but reverses in OPT.}
\label{tab:arch}
\end{table}

\end{document}